\pdfoutput=1

\documentclass[a4paper, 10pt, conference]{ieeeconf}      

\IEEEoverridecommandlockouts                              

\usepackage{cite}

\usepackage{blindtext,graphicx,booktabs,amsmath,amssymb,tabularx,tabu,mathrsfs}
\usepackage[ruled]{algorithm2e}
\usepackage[utf8]{inputenc}
\usepackage{balance}
\newcommand{\frenet}{Frenét }

\title{\LARGE \bf
 Long-Tail Prediction Uncertainty Aware Trajectory Planning for Self-driving Vehicles
}

\author{Weitao Zhou\textsuperscript{1}, Zhong Cao\textsuperscript{1}, Yunkang Xu\textsuperscript{2} \\ Nanshan Deng\textsuperscript{1}, Xiaoyu Liu\textsuperscript{1}, Kun Jiang\textsuperscript{1} and Diange Yang\textsuperscript{1}
\thanks{\textsuperscript{1} School of Vehicle and Mobility, Tsinghua University, Beijing, China. \textit{Email: zwt19@mails.tsinghua.edu.cn, caozhong@tsinghua.edu.cn, \{dns18, xiaoyu-l21\}@mails.tsinghua.edu.cn,  \{jiangkun, ydg\}@tsinghua.edu.cn}, respectively.}
\thanks{\textsuperscript{2} Autonomous Driving \& Advanced Safety System Development Div., Toyota Motor Technical Research And Service (Shanghai) CO.,Ltd. Shanghai, China. \textit{Email: xuyunkang@ttrssh.com.cn}}
\thanks{Corresponding Authors: Z. Cao, K. Jiang and D. Yang.}
}
\makeatletter
\def\endthebibliography{%
  \def\@noitemerr{\@latex@warning{Empty `thebibliography' environment}}%
  \endlist
}
\makeatother

\begin{document}

\maketitle
\thispagestyle{empty}
\pagestyle{empty}

\begin{abstract}
A typical trajectory planner of autonomous driving commonly relies on predicting the future behavior of surrounding obstacles. 
Recently, deep learning technology has been widely adopted to design prediction models due to their impressive performance.
However, such models may fail in the ``long-tail" driving cases where the training data is sparse or unavailable, leading to planner failures. 
To this end, this work proposes a trajectory planner to consider the prediction model uncertainty arising from insufficient data for safer performance.
Firstly, an ensemble network structure estimates the prediction model's uncertainty due to insufficient training data. Then a trajectory planner is designed to consider the worst-case arising from prediction uncertainty. The results show that the proposed method can improve the safety of trajectory planning under the prediction uncertainty caused by insufficient data. At the same time, with sufficient data, the framework will not lead to overly conservative results. This technology helps to improve the safety and reliability of autonomous vehicles under the long-tail data distribution of the real world.


\end{abstract}


\IEEEpeerreviewmaketitle

\section{Introduction}

Self-driving technology is believed to improve traffic efficiency and safety in the future \cite{yang2018intelligent}. A typical self-driving system can be divided into several modules: perception, planning, control.
Among them, the task of the trajectory planner is to compute a feasible trajectory for self-driving vehicles (SDVs) using the estimated status of surrounding objects (provided by the perception module). Then the control module will drive the SDV following the output trajectory.

A trajectory planner commonly relies on the future prediction of surrounding objects in finding the optimal trajectory \cite{schwarting2018planning} \cite{zhou2020integrating}.  
Such prediction task is typically accomplished by a deep learning-based model trained with collected real-world driving data  \cite{mozaffari2020deep}. 
However, the real-world scenarios are usually ``long-tail" distributed  \cite{zhang2021deep}, leading to low model performance in the data-sparse cases  \cite{yang2021delving} \cite{makansi2021exposing}. 
It is because the model ``lacks knowledge" about the environment due to insufficient data, also described as high model uncertainty \cite{hullermeier2021aleatoric} \cite{cao2021confidence}. As a result, the downstream trajectory planner may make risky decisions in ``long-tail" cases.






\begin{figure}
    \centering
    \includegraphics[width=0.5\textwidth]{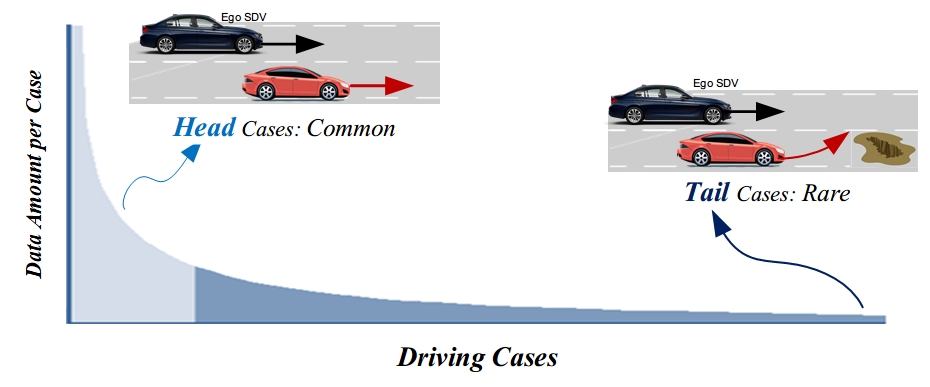}
    \caption{The distribution of real-world driving data. There exists ``long-tail" of driving cases where the amount of data is small.}
    \label{problem}
\end{figure}

A straightforward solution to alleviate the problem is to collect more data on ``long-tail" cases for performance improvement \cite{cao2022trustworthy}. However, it would be too costly as re-encountering a rare driving case can require billions of kilometers of natural driving \cite{wachenfeld2016release}. Furthermore, the prediction model and planner will operate under a known risk of failure until sufficient data is collected. Other existing methods like data re-balancing and data-augmentation \cite{zhang2021deep} are proved effective but still cannot guarantee model performance in ``long-tail" cases.

To this end, we aim to design a planner to work safely under prediction model uncertainty caused by insufficient training data. Related works to improve planning performance on prediction uncertainty can be divided into three methods: 1) Fixed uncertainty bound,  2) Prediction results correction, and 3) Probabilistic prediction results.

\begin{figure*}[ht]
	\centering
	\includegraphics[width=\linewidth]{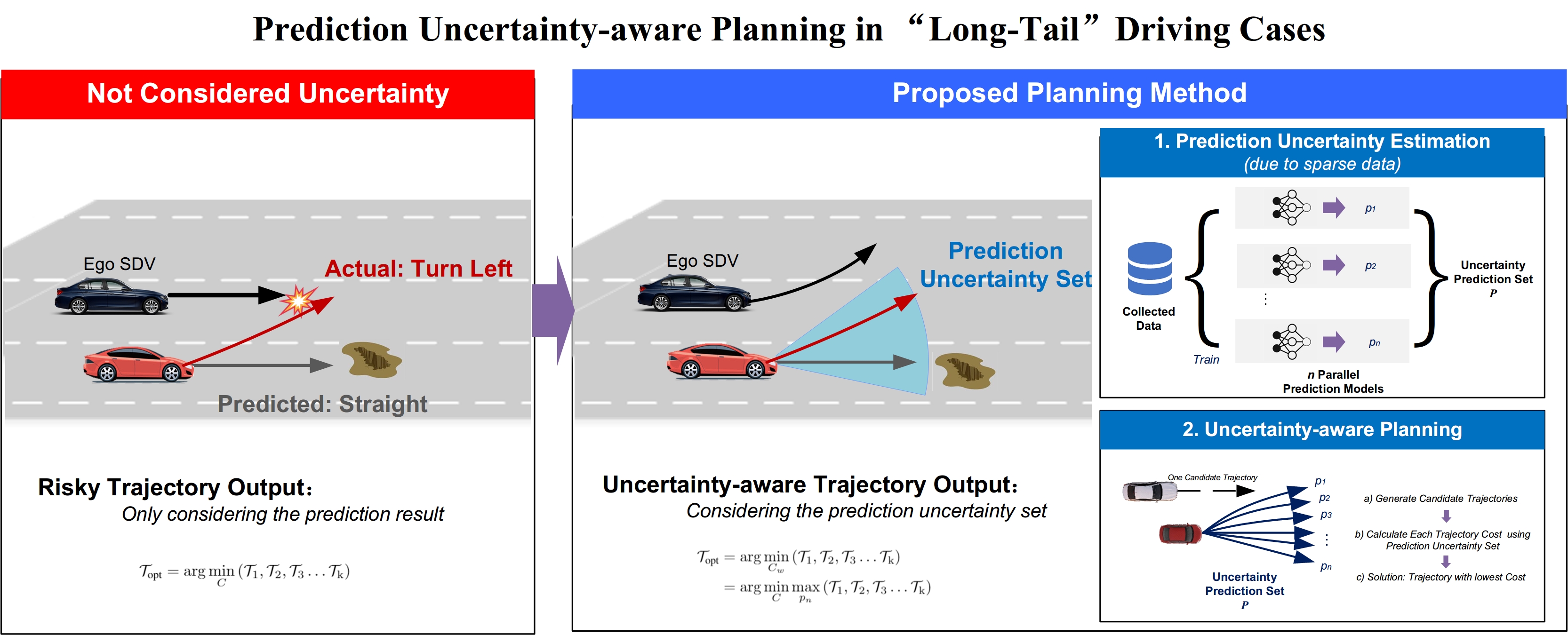}
	\caption{In ``long-tail" driving cases, typical trajectory planners directly use prediction results without considering uncertainty arising from insufficient data , thus producing risky planning results. The proposed planning method first estimates such uncertainty using ensemble methods: train several random-initialized prediction models independently. Then an uncertainty-aware planner will output safe planning results by considering the prediction uncertainty arising in ``long-tail" driving cases.}
	\label{fig:framework}
\end{figure*}

The fixed uncertainty bound method enlarges surrounding objects' bounding boxes to avoid potential prediction uncertainty. For example, the space occupied by obstacles gets larger when the prediction time is longer \cite{xu2014motion}. Furthermore, errors from sensors and localization are taken into account to help determine the extent of enlargement \cite{khaitan2021safe} \cite{cao2021lidar}. The reachable-set planners avoid all possible future of surrounding objects for safety \cite{althoff2021set}.
However, those methods usually rely on a fixed parameter that works in all driving cases. As a result, the AV might be over-conservative even with good prediction results and still risky when the prediction result exceeds the error bound.

The prediction result correction methods incorporate some physical/rule-based models to restrict the prediction results. For example,  physics model of vehicle or pedestrian  \cite{salzmann2020trajectron++}. The driver intention is also considered for modification \cite{zhao2020tnt} \cite{mersch2021maneuver}.
This prior knowledge helps to improve the accuracy of the prediction results overall, but the uncertainty caused by the insufficient data in the ``long-tail'' cases is still a problem.

The probabilistic prediction methods model and consider prediction uncertainty for safer planning performance. The planner takes into account the probabilistic prediction results for safer performance \cite{hubmann2018automated} \cite{khaitan2021safe}. Such probabilistic results can be obtained from pre-modeled distribution \cite{chai2019multipath}, probabilistic learning-based models  \cite{salzmann2020trajectron++}, etc. 
These methods achieve safer performance by considering the prediction's uncertainty. However, the probabilities output by learning-based models are also learned from a ``long-tail" distributed dataset. Thus the wrong prediction probability may still happen in data-sparse cases and cause planning failures.

Unlike the previous methods, this work proposed a planning framework to consider prediction uncertainty arising from the ``long-tail'' data distribution. Such uncertainty is related to data amount and varies in different driving cases. Thus, the pre-design models/rules are challenging to handle. This work will estimate the prediction uncertainty using collected training data and design a planning framework to account for the estimated uncertainty for safer performance.

The contributions include:

1) A method to estimate prediction model uncertainty due to insufficient training data in different driving cases from collected training data.

2) A planning framework considers the estimated prediction uncertainty for safer performance.

The rest of the paper is structured as follows: Section II introduces the preliminaries and problem definition. Section III presents the proposed planning method. Section IV evaluates the algorithm's performance, and Section V summarizes the work with a brief discussion of future work.

\section{Preliminaries and Problem Definitions}

We first define some basic notions.  In this work, a snapshot at time $t$ of the environment's status is defined as a state $s^{(t)}$. Each state is defined including status of ego SDV $s_e^{(t)}$ and $i$ surrounding vehicles $s_i^{(t)}$  . 
\begin{equation}
s^{(t)}=\{s_e^{(t)}, s_1^{(t)}, s_2^{(t)},...s_i^{(t)}\}
\end{equation}
The current state $s_i^{(t)}$ of an agent with its histories for the previous $H$ timesteps is denoted as $x_i^{(t)}=s_i^{(t-H)...(t)}$. A ``driving case" is defined as:
\begin{equation}
x^{(t)} = \{x_e^{(t)}, x_1^{(t)}, x_2^{(t)}, ... x_i^{(t)}\}
\end{equation}
Similarly, the distribution over all surrounding agents' future states for the next $T$ timesteps is defined as:
\begin{equation}
\begin{array}{l}
y^{(t)} = \{y_1^{(t)}, y_2^{(t)}, ..., y_i^{(t)} \} \\
y_i^{(t)} = s_i^{(t)...(t+T)}
\end{array}
\end{equation}
This work aims to solve the following problem: an SDV should plan a safe trajectory under prediction uncertainty caused by ``long-tail'' training data distribution. A typical planning process in a driving case can be described as:

\begin{equation}
\mathcal{T} = f(y^{(t)}, x_e)
\\
\label{planning_all}
\end{equation}
where $\mathcal{T}$ denotes the generated trajectory. The trajectory planner $f$ is driving the SDV using the prediction results $y^{(t)}$ as well as the ego SDV information $x_e$. The prediction results $y^{(t)}$ is estimated from a prediction model $p_{\theta}$ that takes $x^{(t)}$ as input. 
\begin{equation}
\hat{y^{(t)}} = p_{\theta}(x^{(t)})
\end{equation}
In this work, the prediction model $p_{\theta}$ is a deep learning-based model with parameters $\theta$  learned from collected driving data $\theta \leftarrow D$ . 

The ``long-tail" challenge of the described planning process is that the trained model $p_{\theta}$  might be far from optimal in the data-sparse driving cases, leading to a large error of $y^{(t)}$ then sub-optimal or risky $\mathcal{T}$. 

This work will first estimate the model uncertainty $U(p_{\theta}(x^{(t)}))$ to know ``how bad the performance of $p_{\theta}$ is in the current driving case $x^{(t)}$". Then we will design a trajectory planner $f_s$ to consider the estimated uncertainty for more safety performance:

\begin{equation}
\mathcal{T}_s = f_s(U(p_{\theta}(x^{(t)})), x_e)
\\
\label{planning_robust}
\end{equation}
In addition, the specific implementation and type of the prediction model will not be restricted in this work, for example,  LSTM \cite{hou2019interactive}, Graph neural network \cite{gao2020vectornet}. Moreover, this work will not improve the accuracy of the prediction model. We also assume that a control module uses PID or preview control to track the desired trajectory with an acceptable bounded tracking error \cite{xu2021system}.

\section{Methods}
\subsection{Framework}

The framework of the proposed prediction-uncertainty-aware planner is shown in Fig. \ref{fig:framework}. The algorithm includes two main parts: 1) an uncertainty estimator for prediction model outputs in different driving cases. 2) a planning framework designed to account for the uncertainty for safer performance.

\subsection{Prediction Model Uncertainty Estimation}

This section describes the method of prediction uncertainty estimation for deep learning-based models trained on a ``long-tail" distributed dataset. Commonly, a prediction model is trained to give a maximum likelihood estimate $\hat{y^{(t)}} = p_{\theta}(x^{(t)})$ of ground truth ${y^{(t)}}$  \cite{salzmann2020trajectron++} \cite{gao2020vectornet}.  However, the training dataset $D$ has a ``long-tail" of rare driving cases in our problem setting. It means that the data amount of these specific driving cases $N(x^{(t)})$ is low. From a deep learning perspective, the prediction model will be considered to have high epistemic uncertainty under such inputs \cite{hullermeier2021aleatoric}. Furthermore, previous works have also shown that a deep learning-based model has a higher prediction error in these ``long-tail" cases (compared to the cases with more data) \cite{yang2021delving}.

To quantify such uncertainty in ``long-tail" driving cases, we design an ensemble network structure to give a distribution over the estimated prediction results $\hat{y^{(t)}}$, as shown in Fig.  \ref{fig:framework} . The ensemble method efficiently captures the uncertainty of deep learning models without modifications to the original network structure \cite{ganaie2021ensemble}.
Here, we assume a learning-based prediction model $ p_{\theta}$ exists. Then the ensemble network will contain $n$ prediction models $p_{\theta_1}, p_{\theta_2}, ... p_{\theta_n}$ in parallel, each of which has the same structure with  $p_{\theta}$. To train the ensemble network, each prediction network contained will be initialized randomly. 
Then, we can obtain a set of $n$ prediction models that indicate the uncertainty arising from ``long-tail" data distribution. The $n$ prediction model is defined as a prediction model uncertainty set:

\begin{equation}
P= \{p_{\theta_1}, p_{\theta_2}, ... p_{\theta_n}\}
\label{uncertainty_set}
\end{equation}


We adopt a GNN-based prediction model as $p_{\theta}$ in this work. Such type of model is widely used because of its state-of-art performance. 
The chosen model has a similar structure with \cite{salzmann2020trajectron++}. More specific, the model takes historically observed states (e.g., position, velocity) of surrounding obstacles as input.
It contains a self-attention layer to capture the interaction between agents. Then a decoder outputs each object's future probability distribution trajectory. 
Note that the chosen GNN-based prediction model is only an example. The proposed uncertainty estimation and planning method is not limited to a specific model structure. Other learning-based prediction models are also working but were not explored in this work.

\begin{algorithm}  
\caption{Prediction Model Uncertainty Estimation}  
\LinesNumbered  
\KwIn{Training Dataset $D$, Prediction Model $p_{\theta}$}  
\KwOut{A Set of Prediction Model Heads  $p_{\theta_1}, p_{\theta_2}, ... p_{\theta_n}$ }  

Build ensemble network included  $p_{\theta_1}, p_{\theta_2}, ... p_{\theta_n}$  with prediction model $p_{\theta}$

Train each model in ensemble network with dataset $D$ independently

\end{algorithm}

\subsection{Prediction Model Uncertainty-aware Trajectory Planner}

In this section, a planning framework that accounts for the estimated prediction model uncertainty is designed. 
The idea is to consider the worst-case of given prediction uncertainty set instead of a single prediction model result. 

\begin{figure}
    \centering
    \includegraphics[width=0.3\textwidth]{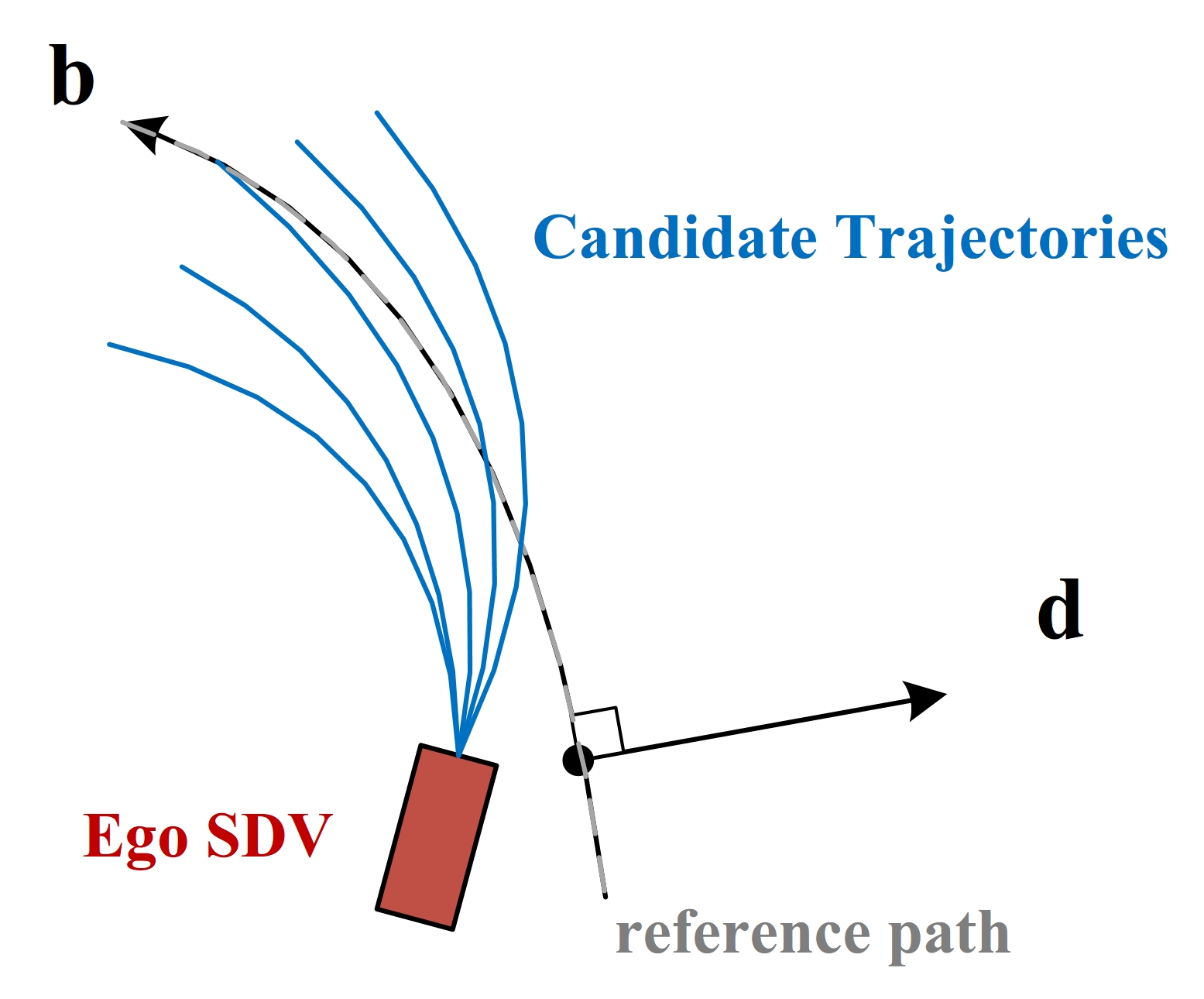}
    \caption{The \frenet frame built on reference path (gray). The red block denotes the ego SDV. The blue curves are the generated candidate trajectories of the lattice planner. The candidate trajectories are various in end states with different lateral and longitudinal distances under \frenet frame.}
    \label{frenet_lattice}
\end{figure}

The proposed planner is based on lattice planner \cite{werling2010optimal}. The lattice planner works on \frenet frame built on each lane/reference path, as shown in Fig. \ref{frenet_lattice}.
The idea of lattice planner is first to sample a group of candidate trajectories, then choose the optimal solution using a pre-defined cost function $C$. In this work, we keep the candidate trajectory generation part and redesign the second part to account for prediction uncertainty for safer performance. 

In first step, the planner will first sample a certain number of trajectory end states $\tau_1, \tau_2, ... \tau_k$. Those end states vary in lateral and longitudinal distances under the \frenet frame. Then a candidate trajectory will be generated using a quintic polynomial with ego vehicle state $x_e^{(t)}$ as start and $\tau_k$ as the end state, as shown in Fig. \ref{frenet_lattice}. In \cite{werling2010optimal}, the authors proved that such a quintic polynomial is the optimal solution to the minimization problem of the cost function:

\begin{equation}
C=k_{j} J_{t}+k_{t} g(T)+k_{p} h\left(b\right)
\label{equ:cost}
\end{equation}
where $J_t$ denotes the time integral of the square of jerk, $T$ denotes the expected time consume from start state to end state, $b$ denotes the lateral offset of end state , $g$ and $h$ are arbitrary functions and  $k_{j}, k_{t}, k_{p}>0$. As a result, polynomials for different end states are grouped as candidate trajectories $\left(\mathcal{T}_{1}, \mathcal{T}_{2}, \mathcal{T}_{3} \ldots \mathcal{T}_{\mathrm{k}}\right)$.

In the original lattice planner, the final solution is obtained by choosing the trajectory with the lowest cost as well as passing the collision checking:

\begin{equation}
\mathcal{T}_{\text {opt}}=\arg \min _{C}\left(\mathcal{T}_{1}, \mathcal{T}_{2}, \mathcal{T}_{3} \ldots \mathcal{T}_{\mathrm{k}}\right) 
\label{equ:cost1}
\end{equation}
\begin{equation}
C(\mathcal{T}_{\mathrm{k}}, p_{\theta})=k_{j} J_{t}+k_{t} g(T)+k_{p} h\left(b\right)+\phi(\mathcal{T}_{\mathrm{k}}, p_{\theta}((x^{(t)})))
\label{equ:cost2}
\end{equation}

\begin{equation}
\phi(\mathcal{T}_{\mathrm{k}}, p_{\theta}((x^{(t)})))= 
\left\{
    \begin{aligned}
    +\infty & , & collision, \\
    0 & , & else.
    \end{aligned}
\right.
\label{collisioncheck}
\end{equation}
The safety performance is guaranteed by collision checking (Eq.  \ref{collisioncheck}), which uses the prediction model results  $p_{\theta}((x^{(t)}))$ directly. The candidate trajectory $\mathcal{T}_{\mathrm{k}}$ and the prediction results of surrounding objects are checked sequentially in time to see if they will collide in the future. As mentioned above, the prediction results might be wrong in ``long-tail" driving cases. Thus the chosen trajectory (that passed the collision checking) might lead to accidents. 

To account for the prediction model uncertainty for safer performance, we redesign the method of choosing the optimal solution from candidate trajectories:

\begin{equation}
\begin{aligned}
&C_w(\mathcal{T}_{\mathrm{k}}, p_{\theta_n}) \\
&=\arg  \max_{p_{\theta_n}  \in P} k_{j} J_{t}+k_{t} g(T)+k_{p} \left(b\right)+\phi(\mathcal{T}_{\mathrm{k}}, p_{\theta_n}((x^{(t)})))
\end{aligned}
\label{cost_worst_case}
\end{equation}

\begin{equation}
\begin{aligned}
\mathcal{T}_{\text {s}}&=\arg \min _{C_w} \left(\mathcal{T}_{1}, \mathcal{T}_{2}, \mathcal{T}_{3} \ldots \mathcal{T}_{\mathrm{k}}\right) \\
&=\arg \min _{C} \max_{p_n} \left(\mathcal{T}_{1}, \mathcal{T}_{2}, \mathcal{T}_{3} \ldots \mathcal{T}_{\mathrm{k}}\right) \\
\label{equ:cost3}
\end{aligned}
\end{equation}
The idea of Eq. \ref{cost_worst_case} is to consider the worst-case in the prediction uncertainty set (described in Eq. \ref{uncertainty_set}). More specific, general practice for trajectory planners is to directly use prediction results as Eq.  \ref{equ:cost1}. This work further considers the uncertainty arising from insufficient training data in ``long-tail" cases. The ``max" in Eq. \ref{equ:cost3} considers the worst-case in the estimated prediction uncertainty. It is easy to find that when $n = 1$, our method is equal to the original lattice planner (Eq.  \ref{equ:cost1}).

In this way, the safe performance of the trajectory planner can be improved and not lead to unnecessarily over-conservative behaviors because 1) When the AV is driving in ``long-tail" cases with high prediction uncertainty, directly using the (maybe significantly wrong) prediction results might lead to a collision. Our method considers the prediction model uncertainty through the prediction model set. The variance of $p_{\theta_1}, p_{\theta_2}, ... p_{\theta_n}$ is expected to be high in these cases and more likely to cover the true value than a single estimation ($n=1$ as the original planner). Thus the planner considered ``worst-case" more likely to consider the risk brought by the true value and output a safe trajectory conservatively.   2) When the AV is driving in driving cases with enough training data, the uncertainty of the prediction model is expected to be low, reflected in closed outputs of $p_{\theta_1}, p_{\theta_2}, ... p_{\theta_n}$. In this case, the ``worst-case" estimated from $n$ prediction results is similar to one single model (baseline). Thus, the proposed planner would not lead to over-conservative trajectories because it works similar to Eq. \ref{equ:cost1}.

The proposed method is based on but not limited to the lattice planer. The idea of considering the ``worst-case'' from the estimated prediction uncertainty set can be applied to other planners that optimize a specific cost function using the prediction results. The way to apply the ``worst-case'' idea is straightforward as from Eq. \ref{equ:cost2} to Eq. \ref{cost_worst_case}.

\begin{algorithm}  
\caption{Safe Planning under Prediction Uncertainty Set}  
\LinesNumbered  
\KwIn{Ego Vehicle State $x_e^{(t)}$, Surrounding Objects State $, x_1^{(t)}, x_2^{(t)}, ... x_i^{(t)}$, Prediction Model Set  $p_{\theta_1}, p_{\theta_2}, ... p_{\theta_n}$ }  
\KwOut{Safe Planning Trajectory $\mathcal{T}_{s}$ }  

Sample candidate trajectories with lattice planner $\mathcal{T}_{1}, \mathcal{T}_{2}, \mathcal{T}_{3} \ldots \mathcal{T}_{\mathrm{k}} \leftarrow x_e$ 

\For{ $\mathcal{T}_{k}$ in $\mathcal{T}_{1}, \mathcal{T}_{2}, \mathcal{T}_{3} \ldots \mathcal{T}_{\mathrm{k}} $}
    {
    Calculated cost  $C_w$ of $\mathcal{T}_{k}$ using Eq. \ref{cost_worst_case} 
            
    }

Find $\mathcal{T}_{s}$ with lowest cost $C_w$ in all trajectories using Eq. \ref{equ:cost3}

\end{algorithm}  

\section{Experiment}
\subsection{Environment Setting} 

\begin{figure}
    \centering
    \includegraphics[width=0.4\textwidth]{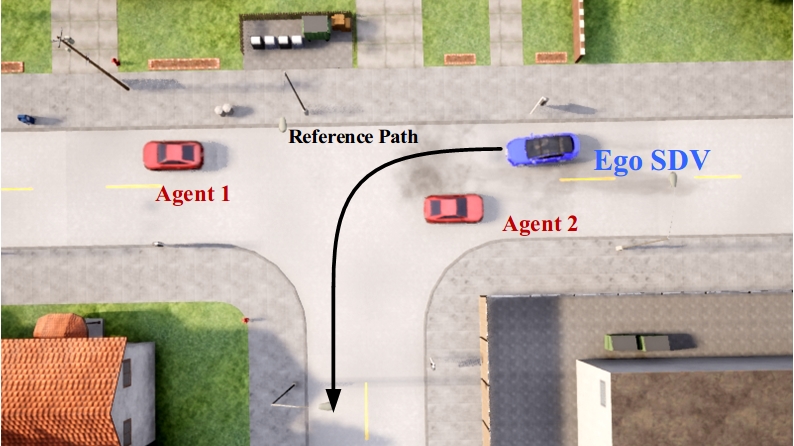}
    \caption{Test environment in CARLA. The ego SDV (blue) takes the black line as the reference path and aims to complete an unprotected left turn task. The surrounding agents (red) are set with random intentions and born positions to generate natural and diverse test cases. The number of agents is also random.}
    \label{carla}
\end{figure}

To train and verify the algorithm, we set up an unprotected left-turn scenario in the CARLA simulator \cite{dosovitskiy2017carla}, as shown in Fig. \ref{carla}.  

The goal of the test SDV is to complete the left-turn task safely considering the surrounding vehicles. The surrounding vehicles going through the intersection will drive without considering traffic lights. Those vehicles are randomly generated with random driving intentions and born positions to create natural and diverse test cases. The simulator provides the dynamics and automatic driving rules. 

\begin{figure}
    \centering
    \includegraphics[width=0.45\textwidth]{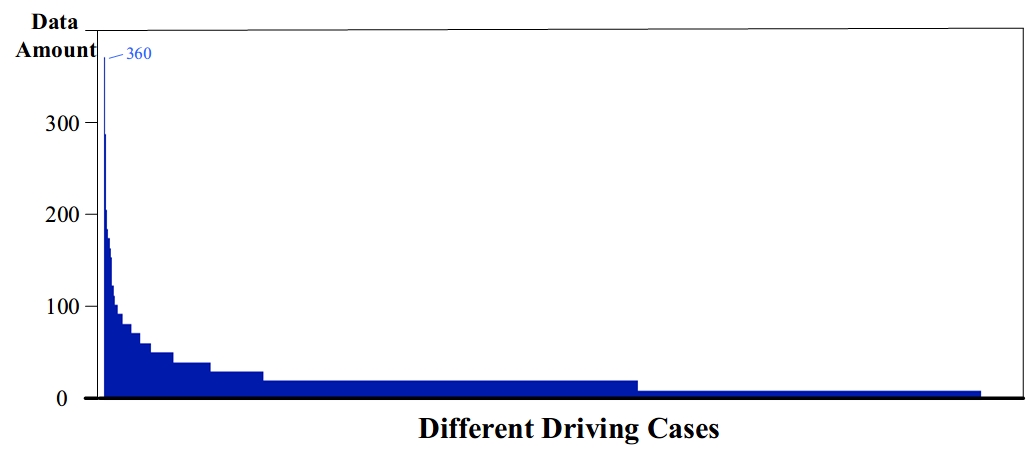}
    \caption{The ``long-tail" data distribution of generated test samples of the unprotected left-turn scenario. The vertical axis is the count for each driving case. The data will also be used to train the prediction model.}
    \label{data_distribution}
\end{figure}

Under this setting, we count the distribution of the generated test samples. In the left-turn scenario, the distribution of driving cases we collected is similar to reality and has a ``long-tail", as shown in Fig. \ref{data_distribution}. The collected data will be used for training the prediction model.

\subsection{Evaluation Metrics} 
We use the following metrics to evaluate the performance of our system:

\subsubsection{Prediction Error Decrease Rate}
This metric measures how much the prediction model uncertainty set results are likely to cover the ground truth. We use two popular metrics in the agent prediction area:

a. Average Displacement Error (ADE): Mean $\ell_{2}$ distance between the ground truth and predicted trajectories.

b. Final Displacement Error (FDE): $\ell_{2}$ distance between the predicted final position and the ground truth final position at the prediction horizon $T$.

The decrease rate of ADE (FDE) is defined as $D_{ADE}(D_{FDE})$ to quantify how much the ensemble prediction models help to decrease the prediction error by covering the ground truth. In our experiment, we will calculated the minimum ADE (FDE) value of all prediction models  $p_{\theta_1}, p_{\theta_2}, ... p_{\theta_n}$  in the prediction model uncertainty set:

\begin{equation}
D_{ADE} =  1-\frac{\arg  \min_{p_{\theta_n}  \in P} ( {ADE(p_{\theta_n})})  } {ADE(p_{\theta_1})}
\end{equation}

\begin{equation}
D_{FDE} = 1-\frac{\arg  \min_{p_{\theta_n}  \in P} ( {FDE(p_{\theta_n})} ) } {FDE(p_{\theta_1})}
\end{equation}

\subsubsection{Safety of Planning Results}
The safety of planning results (under different driving cases) is evaluated as:

\begin{equation}
P_{safe}(x^{(t)}) = \frac{n_{safe}(x^{(t)})}{n_{all}(x^{(t)})}
\end{equation}
where $n_{safe}$ denotes the safe planning times (that not lead to collision), $n_{all}$ denotes the total planning times.

\subsubsection{Efficiency of Planning Results}
The driving efficiency is quantified by two average metrics for planning results of a driving case.

\begin{equation}
P_{ev}(x^{(t)}) =  \frac{\sum_{n=1}^N \overline{v_{\mathcal{T}}}}{N(x^{(t)})}
\end{equation}
where ${N(x^{(t)})}$ denotes how much time the SDV encounters the driving cases in test, $\overline{v_{\mathcal{T}}}$  denote the average expected velocity of a planned trajectory.



\begin{table*}
\caption{Quantitative test results of proposed planning method}
\begin{center}
 \setlength{\tabcolsep}{7mm}{
\begin{tabular}{ccccc}
\toprule  
Number of Prediction Models & $n=1$ (baseline) &$ n=2$ & $n=5$ & $n=10$ \\
\midrule  
Average planning success rate$P_{safe}$  (\%) & 98.09 & 99.23 & 99.61 & 99.65 \\
Average planning velocity $P_{ev}$  (m/s) & 6.89 & 6.66  & 5.91 & 5.63     \\
Average planning velocity $P_{ev}$ (normal cases)  (m/s) & 6.38 & 6.35  & 6.33 & 6.27     \\
Prediction Error Decrease Rate $D_{ADE}$  (\%) &0 & 0.29 & 15.01 & 23.58     \\
Prediction Error Decrease Rate $D_{FDE}$ (\%)  &0 & 2.22 & 15.65 & 23.88         \\

\bottomrule 
\end{tabular}
}
\end{center}
\label{Quantitative}       

\end{table*}

\begin{table}
\caption{Parameters in Proposed Trajectory Planner}
\begin{center}
\begin{tabular}{ccc}
\toprule  
Parameters& Symbol& Value\\
\midrule  
Cost weights of comfort & $k_j$ &0.1 \\
Cost weights of efficiency & $k_t$ & 0.1 \\
Cost weights of offset & $k_p$ & 1.0 \\
Planning time step & $\Delta t$ & 0.1s \\
Sample number& $k$ &  10 \\ 

\bottomrule 
\end{tabular}
\end{center}
\label{parameter}       

\end{table}

\subsection{Baseline}

The baseline used for comparison is the lattice planner \cite{werling2010optimal}. In fact, the proposed planner is equal to the baseline when the number $n$ of prediction models in the ensemble network structure is set to 1. In this case, there is only one prediction model in the ensemble structure, and the worst-case consideration (Eq. \ref{equ:cost3}) is equal to baseline (Eq. \ref{equ:cost1}).

\subsection{Algorithm Parameters and Training} 

We will first train the prediction models during the test using the collected data as shown in Fig. \ref{data_distribution}. Then reconstruct these driving cases to test the proposed planner and baseline ($n=1$). The parameters of the two planners are the same, as shown in the table \ref{parameter}.

\subsection{Computational Performance} 

We implemented our planner on a computer with Intel Xeon(R) E5-2620 CPU, 128Gb RAM, and GeForce GTX 1080 Ti. The planner runs at 10Hz when the number of the prediction model is ten ($n=10$). 
Theoretically, the computational cost of the proposed method is $n$ times of the baseline. Though not explored in this work, the $n$ models can be designed to work in parallel for acceleration in the future. The parallel design will not affect the planning performance. More specific, the $n$ models make predictions and calculate cost (in Eq. \ref{equ:cost}) in parallel. In this way, the proposed method can work with potential larger $n$ numbers and not reduce real-time planning performance too much (compared to baseline).

\begin{figure}
    \centering
    \includegraphics[width=0.45\textwidth]{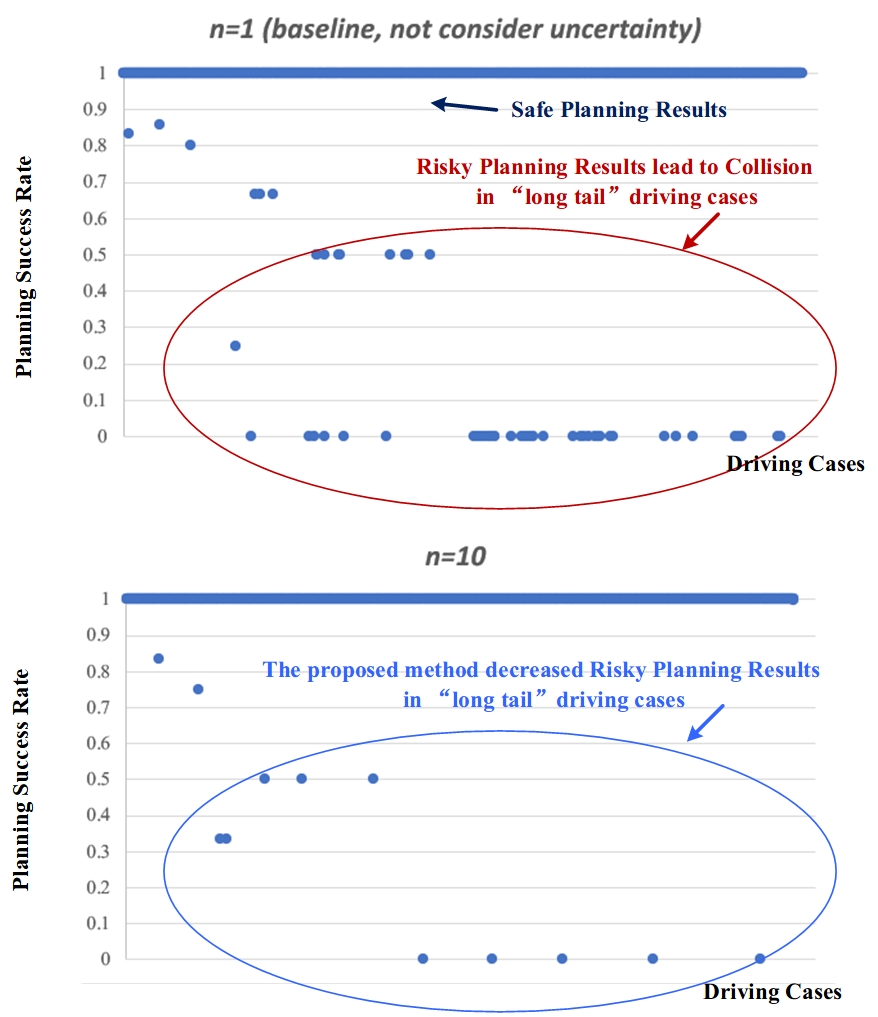}
    \caption{The planning safety of the proposed method and baseline is compared in all driving cases. The horizontal axis represents different driving cases, arranged in the same order as Fig. \ref{data_distribution}, from left to right according to the amount of training data. The vertical axis represents the safety rate $P_{safe}$ of the planning under the driving case. Each point records the performance in a driving case.
The baseline has lower performance in ``long-tail" driving cases due to significant prediction errors (in red circle). The proposed method significantly reduces the number of collisions in these driving cases (blue circle).
}
    \label{planning_safety}
\end{figure}

\subsection{Results} 

\subsubsection{Quantitative results}

We test the proposed uncertainty-aware planner and baseline. The quantitative results are shown in Table. \ref{Quantitative}. Different value of  $n$ (number of prediction models in ensemble model structure) is obtained and tested. Note that the proposed uncertainty-aware planner is equal to baseline when $n$ equals 1. 

The results show that the proposed planner can improve safety performance. The planning success rate in different driving cases increases from 98.09\% to 99.65\%. Fig. \ref{planning_safety} compares the planning safety metrics $P_{safe}$  of proposed method ($n=10$) and baseline in different driving cases. The driving cases in Fig. \ref{planning_safety} are arranged in the same order as Fig. \ref{data_distribution}, which means that the right-hand side is the data-sparse ``long-tail'' cases. The figure shows that the baseline planner not considering prediction uncertainty leads to numerous collisions (leading to a low planning success rate) in ``long-tail'' driving cases. The proposed uncertainty-aware planner significantly reduces the number of risky planning results in ``long-tail'' driving cases.

Different values of $n$ also affect the performance of the proposed uncertainty-aware planner.
The larger $n$ is, the higher the planning success rate is. 
It is because $n$ increases come with a more accurate estimation of prediction uncertainty arising from ``long-tail" data sparsity, leading to safer planning results.
It can be seen that just considering one more prediction uncertainty sampling ($n=2$) can significantly improve planning safety. When $n>5$, the improvement gradually decreases with further increasing $n$ value.

The cost of accounting for prediction uncertainty is reducing the driving efficiency. The average planning velocity will be lower with larger $n$. That is because a larger $n$ means more accurate uncertainty estimates and more coverage of environmental space (in the future), which leads to more conservative decision outputs.

Besides, the efficiency of the proposed planner will not be affected too much in typical driving cases. We define the top 10\% of all cases with the largest amount of data as "non-long-tail cases." The planning efficiency of the proposed method in these cases (6.27 m/s for $n=10$) is close to baseline (6.38 m/s), as shown in Table. \ref{Quantitative}. Thus, the proposed planner would not lead to overly conservative results in ``non-long tail" driving cases.

It also can be seen in Table. \ref{Quantitative} that the prediction error can be decreased with a prediction uncertainty set. A larger $n$ leads to a higher prediction error decrease rate. This is because the outputs from the uncertainty prediction set can be seen as samples from the probability distribution of true value. More samples mean a higher probability of including (or approaching) the truth value. Thus the minimum error in the prediction uncertainty set will be lower (with higher $D_{ADE}, D_{FDE}$). Furthermore, the proposed planner considers the worst-case from the entire prediction uncertainty set. So when the set is more likely to include the true value, the planner will likely plan a safe trajectory that considers the true value.

\begin{figure}
    \centering
    \includegraphics[width=0.5\textwidth]{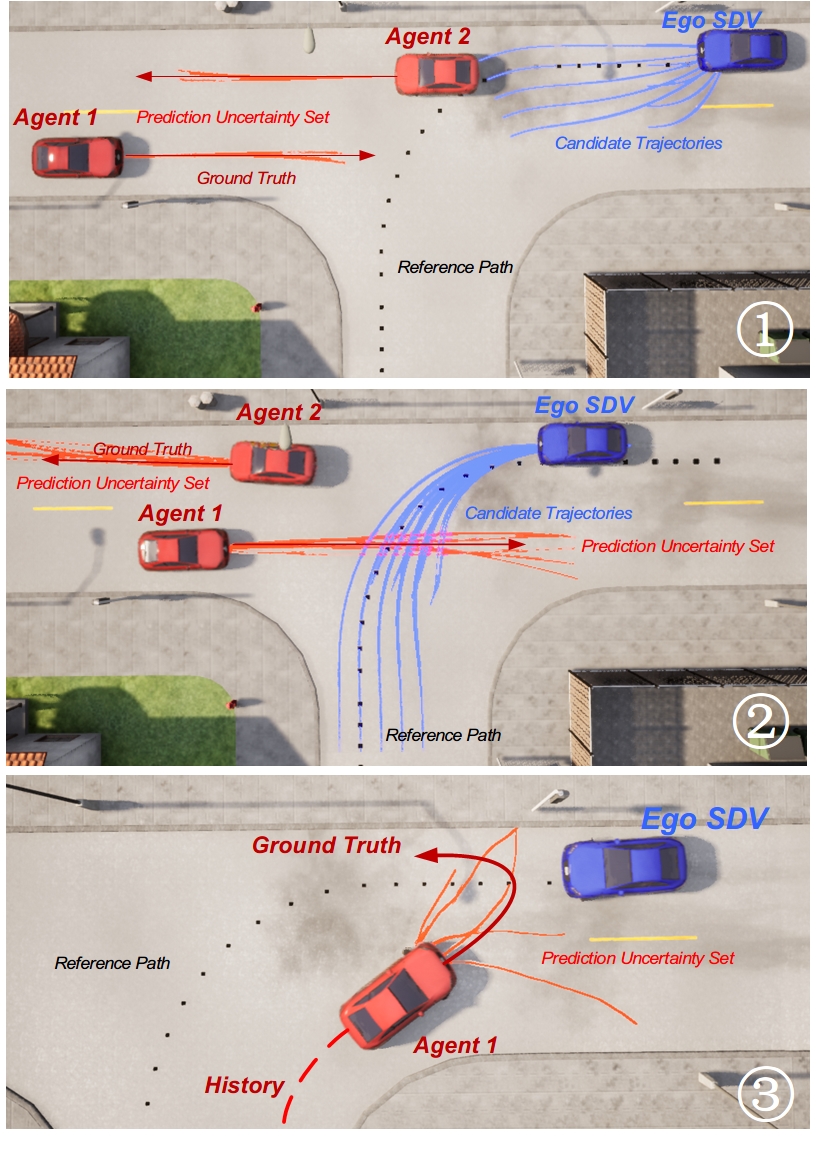}
    \caption{The ``non-long-tail" driving cases (1,2) and ``long-tail" cases (3) recorded during the test. In the figures, the blue curves denotes candidate trajectories of ego SDV $\left(\mathcal{T}_{1}, \mathcal{T}_{2}, \mathcal{T}_{3} \ldots \mathcal{T}_{\mathrm{k}}\right)$, the black points indicates the reference path of unprotected left-turn task, the light red curves are the prediction results from the prediction model uncertainty set $p_{\theta_1}, p_{\theta_2}, ... p_{\theta_n}$, the deep red curves are the ground truth of agents' future trajectories.}
    \label{case}
\end{figure}

\subsubsection{Typical Cases}
 
This section will introduce two example cases to explain how the proposed planner work in ``long tail" and  ``non-long-tail" driving cases. The example is shown in Fig. \ref{case}. 
In the figures, the blue curves denotes candidate trajectories of ego SDV $\left(\mathcal{T}_{1}, \mathcal{T}_{2}, \mathcal{T}_{3} \ldots \mathcal{T}_{\mathrm{k}}\right)$, the light red curves are the prediction results from the prediction model uncertainty set $p_{\theta_1}, p_{\theta_2}, ... p_{\theta_n}$, the deep red curves are the ground truth of agents' future trajectories.

The upper two figures show how the proposed methods behave in  ``non-long-tail" cases where the training data is sufficient. In the two cases, the environmental agents drive to follow a straight line.
The outputs of the prediction uncertainty set are closed around the ground truth. More specific, the output of a single prediction model result ($n=1$, baseline) and results from our proposed prediction uncertainty set are similar with both low errors. Thus, the planning results of the baseline planner (using a single prediction result) and proposed uncertainty-aware planner are similar and both safe. In this case, the proposed planner does not decrease the planning efficiency.

The figure below shows how the proposed method improves planning safety in ``long-tail" driving cases. In this case, the environmental agent makes an unexpected left turn after completing a regular right turn at an intersection. With few training data, a single prediction model (even with probabilistic output) may have high prediction error (for example, predict the vehicle to keep turning right with different speeds), leading to a collision of baseline planner. Here, the prediction uncertainty set gives diverse prediction results, samples from the possible ground truth distribution. As seen from the figure, the results from prediction uncertainty are set to cover most of the ground truth future. Therefore, the uncertainty-aware planner can avoid the potential future of the agent and achieve safe performance.

\section{Conclusion}

In this paper, we proposed a trajectory planning method to consider the prediction model uncertainty arising from the ``long-tail'' distribution of the training dataset. The deep learning-based prediction model may have a high error in data-sparse ``long-tail" driving cases. We defined a notion of prediction model uncertainty to quantify the potential high errors. The model uncertainty is estimated by an ensemble network structure. This network contains several parallel prediction models to output a prediction uncertainty set. We also designed an uncertainty-aware planning method, which considers the worst-case arising from the set for safer performance.

The experiment results show that the proposed planner can significantly improve the safety of SDV in ``long-tail" driving cases where the data is sparse. At the same time, the proposed planner will not lead to
overly conservative results in ``non-long-tail" driving cases where data is sufficient. Besides, the
technology can make the SDV more reliable in real-world ``long-tail" driving cases.

In the future, other methods of estimating deep learning model uncertainty can be employed for better planning performance.

\addtolength{\textheight}{1cm}   
                                  


\section*{ACKNOWLEDGMENT}

This work is supported by the National Natural Science Foundation of China (NSFC) (U1864203, 52102460) and China Postdoctoral Science Foundation (2021M701883)
It is also funded by the Tsinghua University-Toyota Joint Center.
\balance
\appendices

\bibliographystyle{IEEEtran}
\bibliography{ref}

\end{document}